# Predicting the descent into extremism and terrorism

R.O. Lane, W.J. Holmes, C.J. Taylor, H.M. State-Davey, A.J. Wragge

QinetiQ, Malvern Technology Centre, St Andrews Road, Malvern, UK, WR14 3PS

**Abstract**

This paper proposes an approach for automatically analysing and tracking statements in material gathered online and detecting whether the authors of the statements are likely to be involved in extremism or terrorism. The proposed system comprises: online collation of statements that are then encoded in a form amenable to machine learning (ML), an ML component to classify the encoded text, a tracker, and a visualisation system for analysis of results. The detection and tracking concept has been tested using quotes made by terrorists, extremists, campaigners, and politicians, obtained from wikiquote.org. A set of features was extracted for each quote using the state-of-the-art Universal Sentence Encoder (Cer *et al*. 2018), which produces 512-dimensional vectors. The data were used to train and test a support vector machine (SVM) classifier using 10-fold cross-validation. The system was able to correctly detect intentions and attitudes associated with extremism 81% of the time and terrorism 97% of the time, using a dataset of 839 quotes. This accuracy was higher than that which was achieved for a simple baseline system based on n-gram text features. Tracking techniques were also used to perform a temporal analysis of the data, with each quote considered to be a noisy measurement of a person's state of mind. It was demonstrated that the tracking algorithms were able to detect both trends over time and sharp changes in attitude that could be attributed to major events.

## 1. Introduction

Increasingly, malicious activities are carried out online rather than in the physical world. Current intelligence processes dictate that individuals who pose a threat to security are manually identified based on actions or statements they make, before being monitored in more detail. The pool of such people is increasingly diverse, and there are a range of ways they promote propaganda through the internet, using blogs, social media, and videos. Their online and real-world behaviour can be analysed to determine their threat level, so that appropriate action can be taken. However, large data volumes mean that automated processes are needed to assist analysts' understanding of risk and inform the actions of teams that endeavour to prevent harmful behaviour. This paper proposes that analysts will use a language-independent system to track statements made by people over time and determine whether they are likely to become involved in extremism or terrorism. The use of such an automated system will allow analysts to examine a greater volume of data than is currently possible and to prioritise investigations to the most critical individuals.

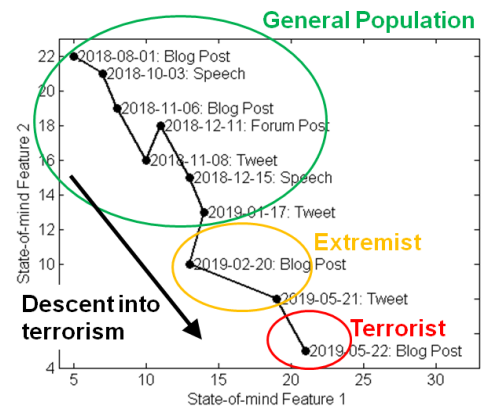

FIGURE 1. System concept for an individual.

The concept of tracking the descent of an individual into extremism and terrorism is illustrated in Figure 1. Each dot on the diagram represents a statement made by the person. The location of dots is a projection of a high-dimensional vector representation of each statement into two dimensions for visualisation purposes (see Section 4 for more detail on encoding text statements as vectors). Different regions of the diagram represent sentiments behind the statements. The individual starts with 'normal' behaviour but over time becomes progressively more extreme and eventually goes on to carry out terrorist attacks.

Use cases for this type of system include:
1. Generating alerts when it looks like someone is likely to become involved in extremism or terrorism (for example after the penultimate statement in Figure 1).
2. Assisting analysts to understand behaviour through visualisation of an individual's state of mind and trends over time. The system would also capture any sudden changes, which could indicate a significant event that has affected the person's attitude.
3. Analysing groups or populations of people by combining their data and determining whether there are general shifts in sentiment.

4.  Analysing the relationships between individuals by looking for correlations in the statements made by people and the time lags involved. If an individual exhibits behavioural changes in a certain direction before other group members then that may indicate the individual is a key influencer. Simultaneous behaviour may indicate a coordinated propaganda campaign.

## 2. System Architecture

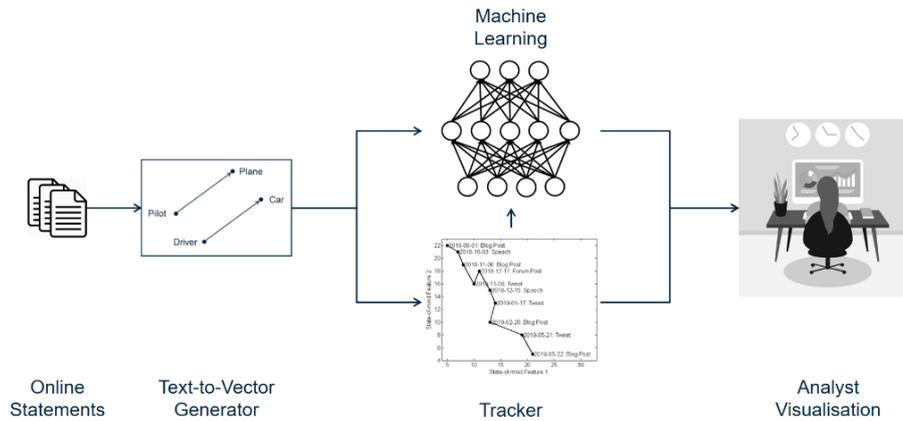

FIGURE 2. System architecture.

The architecture of the proposed system is illustrated in Figure 2. The system comprises: online statement collation, a text-to-vector generator, a machine learning (ML) module, a tracker, and a visualisation system. The vector generator is fed with text data from forum posts, social media, speech transcripts from video (which could in principle be obtained using automatic speech recognition software), or other sources. In the training phase, the ML module uses manually labelled data and the vectors to learn a model of terrorist, extremist, and 'normal' behaviour. In the operational phase, a series of unlabelled statements can then be converted to vectors and tracked using the tracker. The ML module uses the output of the tracker to estimate whether or not the individual is currently or likely to be involved in extremism or terrorism. This information is communicated to the analyst using the visualisation system.

Recently, a number of techniques for converting words or documents to numerical vectors have been published (e.g. Cer *et al.* 2018). These techniques are described in detail in Section 4. Put briefly, words or phrases are represented in vector space in such a way that semantically similar words or phrases are close to each other and relationships between words are captured. For example, the vectors encode relationships like "pilot is to plane as driver is to car" into mathematical equations of the form "pilot - plane = driver - car". Parameters of the vector mapping process are learnt from a large corpus of unlabelled text typical of the language under analysis, which means that the fundamental technique is language independent. The vector representation enables continuous-space techniques, such as clustering and tracking, to be applied to text data.

The vector space can be divided into several regions representing language that is typically associated with various ideologies. Example ideological categories are: general-population ('normal' or centrist), extremist (views could incite violence), and terrorist (willing to carry out acts of violence). For training, a set of statements from the raw text data are manually labelled as belonging to one of these categories. An ML algorithm is trained using the labels and vector representations to assign probabilities of quotes belonging to each category.

In operational use, text from each statement by an individual, whether a blog post, speech or social media post, is modelled as a point in vector space and treated as a noisy measurement of the person's state of mind. Statements can be tracked over time using a predict-update tracking framework over the vector space. At any point in time, the tracker has an estimate of the current state of mind of an individual that will generally be more accurate than an estimate based on any single statement. Future states of mind can be predicted using the past trajectory, with near-term estimates being more accurate than longer term ones. The current or predicted state of mind can be mapped to a probability of being related to a particular ideology using the ML algorithm. If the probability of an individual being potentially involved in terrorism exceeds a threshold an automated alert can be generated. Alerts, combined with the mind-state tracks and visualisation of the semantic vector space, will give analysts actionable insight, allowing them to focus on priority individuals, to understand their behaviour in more detail, and to feed relevant information to teams aiming to prevent harmful behaviour.

## 3. Data Sources

To demonstrate the proof of concept, it was necessary to source a suitable pre-existing data set with favourable licensing conditions and containing statements from a variety of individuals. It was important that each person had made statements over a substantial time period, and that the quotes had already been time-stamped. A suitable source was identified as Wikiquote[1], a "free online compendium of sourced quotations from notable people and creative works in every language and translations of non-English quotes". Wikiquote has pages from 59 active languages that each have more than 100 people. The English language version has quotes from 35,000 people. For the purposes of this study only quotes in English were used but the techniques could be applied to other languages.

Since the ultimate aim is to detect and analyse people from a general population that are likely to become involved in extremism or terrorism, the following groups of people were included for analysis:

- Terrorists (Islamist and Far Right);
- Extremists (Islamist and Far Right);
- UK and EU Politicians;
- Islamic anti-violence campaigners.

Politicians occasionally talk about terrorism or extremism without being involved in it but mostly talk about other topics. For this study they acted as a surrogate for the general population. Islamic anti-violence campaigners have been included to ensure that bias is not inadvertently introduced to the analysis and that people who talk about Islamic topics are not incorrectly categorised. The complete data set comprised 839 quotes from 48 different people.

### 3.1. Data extraction and cleansing

Information on the Wikiquote website is input by the public as free-form text, which is organised into sections, either by decade, year, or speech date. There is typically one bullet point per quote and a sub-bullet point containing the source of the quote and the date it was made. A scraping tool was written to extract relevant information from the semi-structured web page and store this in a structured table that included name, quote text, date of quote, data source, and a serial number to uniquely identify the quote. A difficulty encountered when writing the tool was that formatting of the pages is not enforced, which complicates automatic data extraction. Dates could appear in different places on a page, and the same date could be referred to in various ways, including different day/month orderings ("27/09/2018", "09/27/2018"), two-digit years ("27/09/18"), full or abbreviated month names ("September 27, 2018", "27 Sep 2018"), and having missing years with year implied by a subheading elsewhere ("September 27").

The natural language processing library spaCy was used to extract dates from the text[2]. To reliably detect all date formats in the Wikiquote entries, a custom rule-based date-extraction module was written for spaCy using regular-expression pattern matching. In situations where a date was missing or could not be extracted, the Wikiquote website was updated manually after checking the citation or performing an internet search. Where multiple dates existed or the date was imprecise (e.g. with only a year specified) the earliest possible date was taken, which was indeed correct for most citations where the date of the article was later than the date of the actual quote. These types of formatting issues should not apply in a real monitoring system as that would rely on published application programming interfaces (APIs).

### 3.2. Data labelling

Each quote was assigned to one of three categories:

- c     Centrist: someone with moderate views (general population);
- e     Extremist: someone with vocal opposition to fundamental values including democracy;
- t     Terrorist: criminal with an ideological motive, including willingness to kill or harm civilians.

The labelling was achieved by providing every quote to three independent raters[3] and combining the ratings to obtain a single label for each quote. To avoid bias, the quotes were anonymised and randomised before being provided to the raters, together with a detailed definition of each of the three categories.

Of the 839 quotes, 681 were labelled as 'Centrist', 130 as 'Extremist' and 28 as 'Terrorist'. This set of labelled quotes was used for experiments to investigate automated analysis of potentially extremist or terrorist material.

---

[1] https://www.wikiquote.org

[2] https://spacy.io

[3] One female and two males from the same team produced ratings without discussion with each other.

## 4. Representing text for use with machine learning algorithms

To apply ML algorithms, the text needs to be converted into a form that captures the linguistic content in a way that is amenable to applying mathematical techniques to make assessments of similarity and perform classification. The simplest type of representation has one dimension per word with a unit value for the relevant word and zero for all the other words. Such a "one-hot" representation is, however, limited in that it does not capture syntactic or semantic information (e.g. different word endings for the same verb or different words with similar meanings), or anything about the relationships between words, as well as suffering issues of very high dimensionality and data sparseness, due to the large number of possible words. A variety of Natural Language Processing (NLP) techniques have been developed that aim to provide a numerical representation that captures more useful information about text content.

### 4.1. N-grams

In the context of NLP, an n-gram is a contiguous sequence of n items, such as words from a given sequence of text or speech. If n=1, only individual words ("unigrams") are represented, but for larger values of n, word sequence information is captured. As the value of n is increased, more detailed information can be captured, at the expense of data sparsity. A popular compromise is to use n=2, capturing two-word phrases or "bigrams". Various filtering techniques can be used to help capture the most relevant information in the n-grams. One widely-used technique is to remove 'stop words' (e.g. "the", "is", "at", "on"), which are non-content words that do not provide information about the topic of the text.

N-grams are widely used for NLP tasks including text categorisation (e.g. Rahmoun *et al.* 2007), and are also useful for helping an analyst understand words and phrases that are most indicative of particular concepts. However, simple word-based n-grams do not capture anything about meaning so cannot represent semantic relationships between words.

### 4.2. Text-to-vector embeddings

The desire for a numeric text representation that represents meaning in an efficient way has been addressed by a variety of methods for converting text into a continuous vector space (of typically 200-600 dimensions) that captures relationships between words. The underlying concept is one that comes from the semantic theory of language usage, i.e. words that are used and occur in the same contexts tend to have similar meanings. Hence the aim of the vector representation, often referred to as a word embedding or text embedding, is that semantically similar words will occupy similar regions in vector space. A good word embedding vector can capture relationships between words in a simple way by mathematical operations. For example, (*king - man + woman)* will result in a similar vector to that of *queen*. Therefore tasks such as measuring similarity between pieces of text, identifying sentiment, and classification of text into different topics can take place.

The use of predictive models for deriving word vectors was popularised by Mikolov *et al*. (2013) in their development of word2vec. The concept involves training a neural network to predict words from neighbouring words, and the weights of the trained network provide the feature vector. There are two different algorithms in word2vec: continuous bag of words (CBOW) or continuous skip-gram. In CBOW, the model predicts the current word from the average of a window of surrounding words. In continuous skip-gram, the model uses the current word to predict the surrounding words. As an alternative to a predictive model, Pennington *et al*. (2014) proposed a count-based word co-occurrence model called Global Vectors for Word Representation (GloVe). GloVe produces the embeddings based on a log-bilinear regression model over word pairs using unsupervised learning. Further developments include Fasttext (Bojanowski *et al.* 2017), which is a development of word2vec to represent word vectors as a sum of substrings which can hence accommodate new words, misspellings and other word-form variability. More recent models, such as Embeddings from Language Models (ELMo) proposed by Peters et al. (2018) and Bidirectional Encoder Representations from Transformers (BERT) from Devlin et al. (2019), have made further extensions to generate word vectors that are context specific, allowing different meanings for the same word.

The approaches described above are designed to capture semantic properties of individual words. For applications such as the one being developed for the current study, it is necessary to process variable-length sequences of words. An intuitive way of representing a word sequence is to perform some sort of averaging of the word vectors (e.g. Speer *et al.* 2017), but this approach does not take advantage of word-order information. Methods that encode word order include the Paragraph Vector (also known as doc2vec) proposed by Le and Mikolov (2014) as an extension of their earlier work on word2vec. The algorithm was designed to learn fixed-length vector representations of arbitrary-length blocks of text by adding a paragraph identifier to the inputs previously used for word2vec.

A variety of other sentence-level embedding schemes have been proposed, but a general issue concerns ability to perform well on diverse tasks without requiring large quantities of task-specific labelled training data. This issue was targeted with the release of the Universal Sentence Encoder (Cer *et al.* 2018), which has demonstrated good accuracy on a range of NLP tasks. Two versions are available, both of which use multi-task learning with a variety of labelled and unlabelled training data. The first version uses a neural net transformer architecture to compute context-aware representations of words in a sentence that take into account both the ordering and identity of other words. The context-aware word representations are averaged together to obtain a sentence-level embedding. The transformer version has good accuracy but at the expense of high processing and memory cost for long sentences. The alternative is a Deep Averaging Network (DAN), which computes word and bigram embeddings, averages these and passes them through a deep neural net. DAN is faster than the transformer but tends to be less accurate. In a comparison with GloVe embeddings, Cer *et al.* showed that the Universal Sentence Encoder demonstrated lower levels of bias when measured against sex, age, and names from different regions of the world. This characteristic combined with its robustness to encoding sentences for a wide variety of tasks led to the choice of the Universal Sentence Encoder for the text vector representation used in the current study. While the current work uses only the system for English, there are now models available for a range of languages, including those with different writing systems, such as Arabic, Russian, and Chinese (Yang *et al.* 2019).

## 5. Machine learning experiments

Experiments were carried out to investigate the use of machine learning models to classify the 839 quotes that form the database described in Section 3. As the total quantity of data is quite small, especially for the extremist and terrorist categories, the experimental procedure for the classification experiments used 10-fold cross validation (see for example Arlot & Celisse, 2010), ensuring that each quote contributed exactly once to the overall results while having no overlap between training and test data. All classification experiments used support vector machines (SVMs), which are widely used for classification tasks (see for example Burgess, 1998). An SVM is a discriminative classifier defined by a separating hyperplane. The parameters are determined from labelled training data to find the hyperplane in N-dimensional space (where N is the number of features) that optimises separation of the data points into the available classes. The experiments here used linear and radial basis function (RBF) SVMs. In all cases the parameters of the models were optimised using a grid search and results are quoted for the best configuration in each case. For all experiments an average confusion matrix was computed showing actual label against predicted label, to give information on confusions as well as the proportion of correctly recognised labels.

In the following sections, experiments using N-gram text features are first described to establish a baseline, followed by investigations using vector features derived using the Universal Sentence Encoder.

### 5.1. N-gram text features

Given the small size of the current data set, only unigrams (n=1, single words) and bigrams (n=2, two-word phrases) were extracted, applying stop-word removal in both cases. Stop-word removal has much greater effect for bigrams than for unigrams so after stop-word removal there were 7178 unigrams and 6426 bigrams in the learnt dictionary. To identify the words and phrases that were most indicative of the three categories, a Term Frequency - Inverse Document Frequency (TF-IDF) measure was computed (see for example Ramos, 2003). TF-IDF is a measurement used in text analysis to indicate the importance of a particular term in relation to the corpus. In this case each terrorism category (i.e. "t", "e", and "c") was treated as a document in a corpus to identify the key terms associated with each category. It was found that the words and phrases most associated with the centrist "c" category were mainly related to UK Politics. The words and phrases most associated with the extremist "e" category include cursing and swear terms and references to religious terms and American politics (due to the presence of right-wing American extremists in the data set). The words most associated with the terrorist "t" category made reference to the attacks on the "twin towers" and other buildings and infrastructure that have been attacked by terrorists.

The label composition of the data is very imbalanced: the centrist category accounts for 81% of the data, the extremist category accounts for 15% of the data, and the terrorist category for just 3%. Therefore up-sampling was used, whereby the minority classes were randomly sampled (with replacement) to be the same size as the majority class. These classification experiments used linear SVMs to predict the three-way classification of the terrorism categories based first on unigram and then bigram features, which produced the confusion matrices shown in Figures 3 and 4 respectively. The best overall classification accuracy was obtained using unigram features. It can be seen that the classifier is very good at predicting the centrist ("c") category, but not the other two categories. The confusion matrix shows that both the extremist and terrorist categories will often be confused with the majority centrist category despite the use of up-sampling to mitigate against the imbalance in the number of quotes assigned to each category. Because the complete set of n-gram features is collated across the entire data set, inevitably it is dominated by the most common centrist class. There is also a general problem that there are a very large number of features (n-grams) for a relatively small total quantity of data. There are several options that could be explored for improving on the baseline n-gram accuracy, including techniques for

6 Predicting the descent into extremism and terrorism feature reduction, alternative variants of the SVM, different decision functions, or alternative machine learning approaches that use a sub-set of the features (e.g. a decision tree or random forest, see for example Kowsari *et al*. 2019). For the current study, the use of the alternative text vector features was explored, as described in the next section.

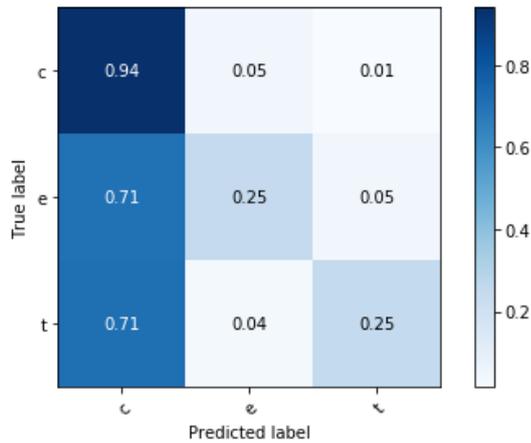 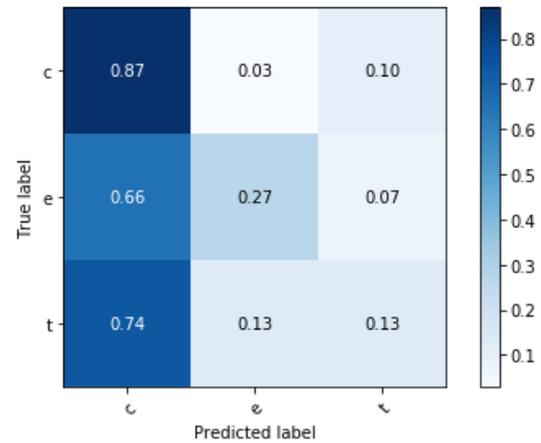

FIGURE 3. Proportion confusion matrix for three-way classification of quotes using unigram features.

FIGURE 4. Proportion confusion matrix for three-way classification of quotes using bigram features.

**5.2. Text vector features**

The aim of the experiments in this section was to investigate the extent to which it should be possible to separate the data into different classes based on the text vectors described in Section 4.2. The experiments started with the 512-element vectors that are output by the universal sentence encoder, and used Principal Components Analysis (PCA) and Linear Discriminant Analysis (LDA) to reduce the dimensionality of the feature set so that it could be visualised graphically. For an overview of PCA and LDA, see for example Kowsari et al. (2019). PCA is a technique for transforming a data set so that the first principal component has the largest possible variance (that is, accounts for as much of the variability in the data as possible), and each succeeding component in turn has the highest variance possible under the constraint that it is orthogonal to (i.e. uncorrelated with) the preceding components. LDA is a related technique that also takes into account the class labels (e.g. "Terrorist") and looks for the transformation that maximises the difference between the classes.

The majority of the variability in the data is captured in the first two components and these can be used to create a two-dimensional plot to give an impression of how the data are distributed. By including markers to indicate data labels, it is possible to see how easily the data fall into different classes. LDA was found to be more useful than PCA for visualising the separation between the classes, and a labelled LDA plot is shown in Figure 5. It can be seen that the small number of quotes labelled as terrorist appear quite distinct from the other quotes. The extremist and centrist quotes also appear as two separate groups, though with more overlap and a less definite boundary between them. From the figure, it appears that it should be possible to distinguish between different classes for much of the data.

Both linear and RBF kernel SVMs were evaluated for classification ability, in combination with PCA to reduce the vector dimensionality. The choice of kernel, parameters of the kernel, and number of PCA components were all optimised by grid search. As there is a high proportion of centrist quotes, a balanced accuracy metric with weights inversely proportional to class frequencies[4] was used for optimisation of the SVM parameters, to equalise the weight given to the extremist and terrorist categories. Although the differences between the results for the linear and RBF SVMs were small, the RBF kernels always outperformed the linear kernels and were therefore selected.

A three-way classifier was trained and tested to give the confusion matrix shown in Figure 6. Although the 76% correct detection of the centrist category is worse than the 94% achieved with the unigram features, the detection of terrorist and extremist categories is much more accurate. Detection of the terrorist category is very high at 97%. Extremism is more difficult to separate out from terrorism, with only 47% correctly classified and 43% being confused with the Terrorist category.

---

[4] To obtain a balanced accuracy metric, the number of correct decisions for each class *y* was multiplied by: (total observation count) / (number of classes * observation count for class *y*)

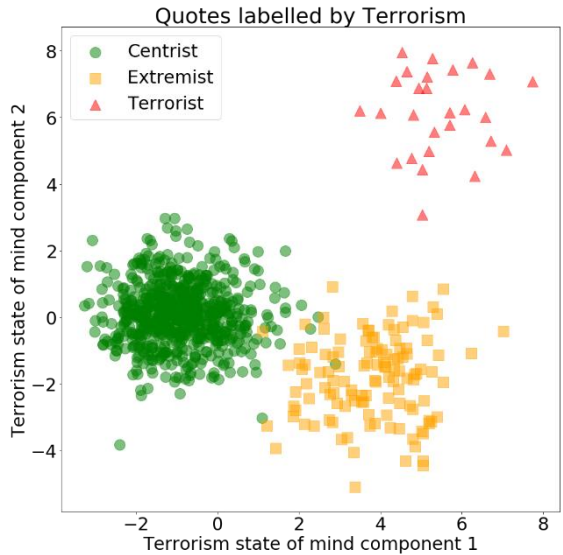

FIGURE 5. Projection of text-vector quote data onto two axes using LDA and terrorism labels.

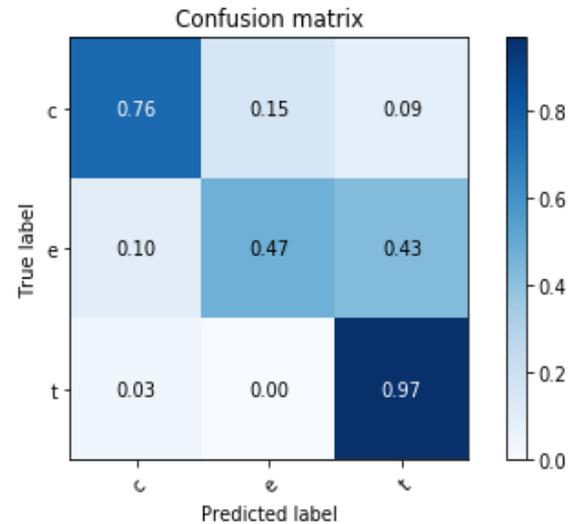

FIGURE 6. Proportion confusion matrix for three-way classification of quotes using text vectors.

Two binary classification tasks were also assessed: detection of terrorism and detection of extremism. For binary classification, in addition to confusion matrices, performance of the SVM was analysed as a detector for the class of interest (e.g. "Terrorist"), by obtaining class membership probabilities as output. By varying a threshold parameter on these probabilities, it was possible to plot a Receiver Operating Characteristic (ROC) curve, which shows the trade-off between true positive rate (sensitivity to the class of interest) and false positive rate (specificity to that class). The area under the curve (AUC) gives a single measure of how well the detector is working, with AUC=1 defining a perfect detector and AUC=0.5 being a detector that randomly chooses between the classes.

The detection of terrorism task examined the ability to separate out the terrorist category from the other two categories, and it can be seen from the confusion matrix in Figure 7 that the ability to detect the terrorist quotes is very good at 97%. The ability to recognise the combined centrist/extremist categories is also very good at 95%. It is interesting that the issues of confusion between the extremist and terrorist categories have been solved by not simultaneously trying to detect terrorists while also trying to distinguish between extremists and centrists. The ROC curve shown in Figure 8 gives more detail on the trade-off between correct detection of terrorist labels and false positive errors. Again it can be seen that the performance is very good, with an AUC of 0.97. Perfect detection of all terrorist labels can be achieved if a false positive rate of 10% can be tolerated.

For early detection of descent into terrorism, it is important not only to be able to detect terrorists, but also to be able to detect early signs of extremist tendencies. Such detection equates to being able to separate out extremist quotes from the neutral centrist ones. Hence this experiment removed all quotes labelled as terrorist and assessed ability to detect the extremist category. The confusion matrix in Figure 9 shows reasonable recall of 81% for correct detection of the extremist quotes. This level of accuracy is however quite a lot lower than the terrorist detection performance of 97%, which suggests that separating out extremists is more difficult than detecting terrorism. The ROC curve shown in Figure 10 gives more detail on the trade-off between correct recognition of extremist labels and false positive errors. Performance can be seen to be good, with an AUC of 0.91, but not as accurate as the terrorist detection performance shown in Figure 8. In the case of extremist labels, perfect detection would involve a high proportion of false positives.

## 6. Tracking and Prediction

While it is of interest to classify either individual quotes or a person's general outlook, based on their complete set of quotes, it is also particularly important to analyse trends over time. These trends can be used to determine whether someone's behaviour is relatively consistent, they are drifting towards extremism or terrorism, or a sudden change in behaviour has occurred. Sudden changes could be in reaction to an event or meeting a new influential person or group of people. Time-based analysis could be extended to groups or whole populations.

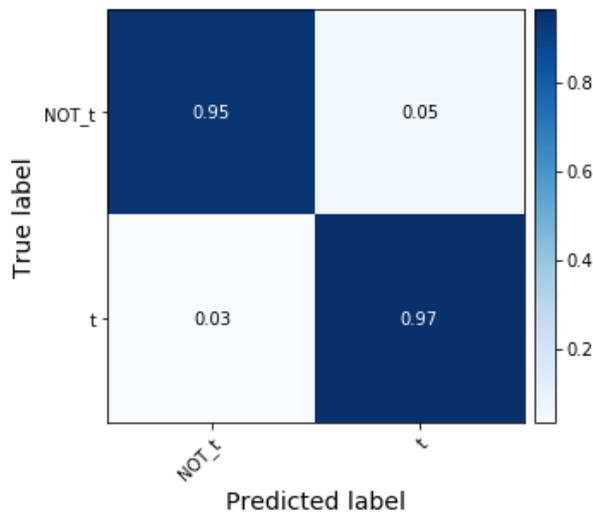

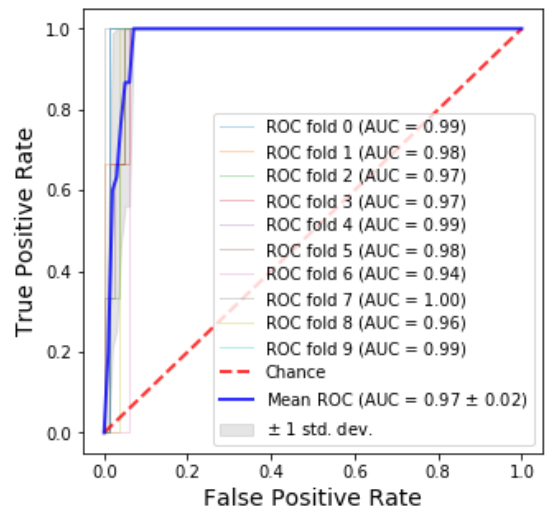

FIGURE 7. Proportion confusion matrix for binary detection of Terrorist quotes from Centrist and Extremist quotes.

FIGURE 8. ROC curve for detection of Terrorist quotes from Centrist and Extremist quotes.

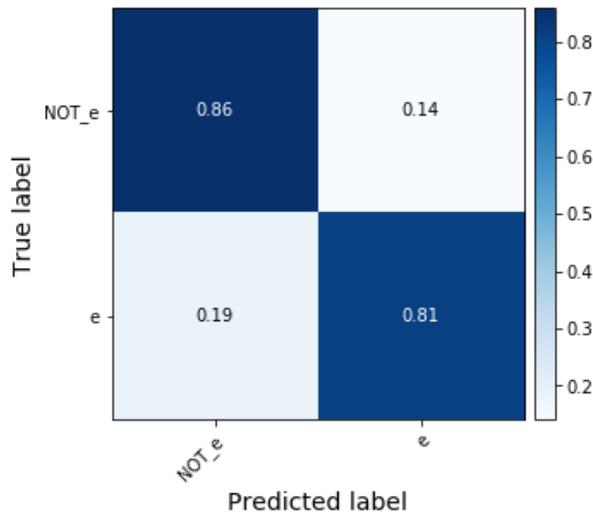

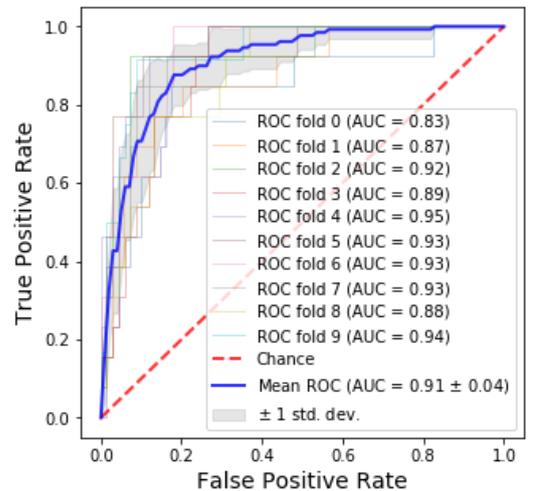

FIGURE 9. Proportion confusion matrix for binary detection of Extremist quotes from Centrist quotes.

FIGURE 10. ROC curve for detection of Extremist quotes from Centrist quotes.

Statements made by individuals can be considered to be a noisy measurement of their state of mind. Using the framework developed over previous sections, this state of mind is represented as a high dimensional vector in the same state space as the vectors produced from the text of their statements. The state of mind can be tracked over time using standard tracking algorithms originally developed for sensor data, such as the Kalman filter (Welch & Bishop, 1995). The generated tracks not only output an estimate of the current state of mind but also its uncertainty. As more data are gathered, confidence in the current estimate and trajectory increases – the tracker estimate is more accurate than using a single statement. Future mind states of the individuals under analysis can be estimated using the current trajectory, with near-term estimates being more accurate than longer-term ones. This provides the capability to predict whether the language being used is becoming more extreme and, consequently, the predicted behaviour is more extreme.

A Kalman filter has been implemented to track the data. While in principle it would be possible to track in 512 dimensions, this introduces issues associated with high-dimensional data. Instead, the data was projected to a two dimensional variable $(x_1, x_2)$ using the LDA technique described in section 5.2. A person's state of mind $\mathbf{x}$ is postulated to change over time $t$ according to a nearly constant velocity model $\mathbf{x}_{t+\Delta T} = \mathbf{F}\mathbf{x}_t + \mathbf{w}$, where $\mathbf{x} = [x_1, x_1', x_2, x_2']^T$. The transition matrix $\mathbf{F}$ and covariance $\mathbf{Q}$ of the zero-mean normally distributed process noise $\mathbf{w} \sim N(\mathbf{0}, \mathbf{Q})$ are as follows (Lane *et al*. 2014).

$$\mathbf{F} = \begin{bmatrix} 1 & \Delta T & 0 & 0 \\ 0 & 1 & 0 & 0 \\ 0 & 0 & 1 & \Delta T \\ 0 & 0 & 0 & 1 \end{bmatrix} \quad (1) \qquad \mathbf{Q} = \begin{bmatrix} \Delta T^3/3 & \Delta T^2/2 & 0 & 0 \\ \Delta T^2/2 & \Delta T & 0 & 0 \\ 0 & 0 & \Delta T^3/3 & \Delta T^2/2 \\ 0 & 0 & \Delta T^2/2 & \Delta T \end{bmatrix} \sigma^2 \quad (2)$$

Each two-dimensional quote vector $\mathbf{z}_t$ is modelled as being a noisy measurement of the underlying state of mind at that time via $\mathbf{z}_t = \mathbf{H}\mathbf{x}_t + \mathbf{v}_t$. The matrix $\mathbf{H}$ selects the $x_1$ and $x_2$ variables from $\mathbf{x}$. The noise vector $\mathbf{v}_t$ has a zero-mean normal distribution with covariance $\mathbf{R_z}$ that depends on the current measurement. The reason for using measurement-dependent noise is as follows. It has been observed that politicians in the data set rarely make extremist statements and never make terrorist ones. Extremists primarily make a mixture of centrist and extremist statements but tend not to make terrorist ones. Terrorists make a mixture of centrist, extremist, and terrorist statements. Based on proportions in the data set, the probability of making a statement of type $s \in \{c, e, t\}$ given the person is in a state-of-mind type $k \in \{c, e, t\}$ is encoded below in the matrix (3) and the proportion of each state-of-mind type is in (4). To compute the proportions, each person in the data set was assigned a single state-of-mind type. The data implies that: a single centrist statement is not a strong indicator of state of mind; an extremist statement indicates someone may be an extremist or terrorist; a terrorist statement strongly indicates that someone is indeed a terrorist. The actual probabilities $p(k|s)$ could in principle also be tabulated.

$$p(s|k) = \begin{bmatrix} s \backslash k & c & e & t \\ c & 0.993 & 0.584 & 0.220 \\ e & 0.007 & 0.409 & 0.532 \\ t & 0 & 0.007 & 0.248 \end{bmatrix} \quad (3) \qquad p(k) = \begin{bmatrix} c & e & t \\ 0.813 & 0.083 & 0.104 \end{bmatrix} \quad (4)$$

The Kalman filter requires knowledge of $p(\mathbf{z}|\mathbf{x})$. This is given by the approximation (5), where $p(\mathbf{x}|k) = N(\mathbf{Hx}|\boldsymbol{\mu}_\mathbf{x}^k, \boldsymbol{\Sigma}_\mathbf{x}^k)$ and $p(\mathbf{z}|k) = N(\mathbf{z}|\boldsymbol{\mu}_\mathbf{z}^k, \boldsymbol{\Sigma}_\mathbf{z}^k)$. The variables $\boldsymbol{\mu}_\mathbf{x}^k, \boldsymbol{\Sigma}_\mathbf{x}^k$ and $\boldsymbol{\mu}_\mathbf{z}^k, \boldsymbol{\Sigma}_\mathbf{z}^k$ are respectively the mean and covariance of the distribution of quote vectors for each statement type (for $\mathbf{x}$) or person type (for $\mathbf{z}$). For use in the Kalman filter, the Gaussian mixture in (5) is approximated in (6) by $N(\mathbf{z}|\mathbf{Hx}, \mathbf{R_z})$, where $\mathbf{R_z}$ is the covariance of (5) computed via mixture reduction.

$$p(\mathbf{z}|\mathbf{x}) \propto p(\mathbf{x}|\mathbf{z}) \approx \sum_k p(\mathbf{x}|k)\, p(\mathbf{z}|k) p(k) \quad (5) \qquad p(\mathbf{z}|\mathbf{x}) \approx N(\mathbf{z}|\mathbf{Hx}, \mathbf{R_z}) \quad (6)$$

The above Kalman filter was applied to quotes made by Osama bin Laden. The process noise variance $\sigma^2$ was set to 0.1, and independent prior distributions $p(x_1) = p(x_2) = N(0, 4^2)$ and $p(x_1') = p(x_2') = N(0, 0.3^2)$ were used. The priors for $x_1, x_2$ were based on the distribution of all quotes for all people. In the absence of ground truth, noise variance $\sigma^2$ and priors for $x_1', x_2'$ were set based on manual data analysis and judgement. The tracker output in terms of a two-dimensional state-of-mind is shown in Figure 11. For visualisation purposes, this is overlaid onto the decision regions of a simple centrist/extremist/terrorist linear classifier trained on the full data set. It is seen that initial quotes made by bin Laden are considered extremist but not terrorist. He then went on to make more terrorist statements for a period of time after 2001 before returning to a less terrorist but still extremist standpoint. This is further illustrated in Figure 12, which shows the second dimension of the state-of-mind as a function of time. The effect of the observation-dependent noise model is clear. Once bin Laden started making some terrorist statements the tracker estimate of this variable remains high, despite him making a greater number less severe statements during the same time period.

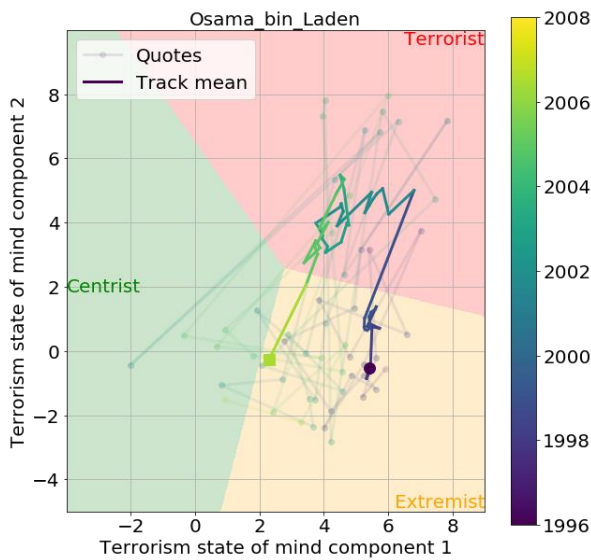

FIGURE 11.   Quote observations and state-of-mind track in two dimensions, overlaid on linear classifier regions.

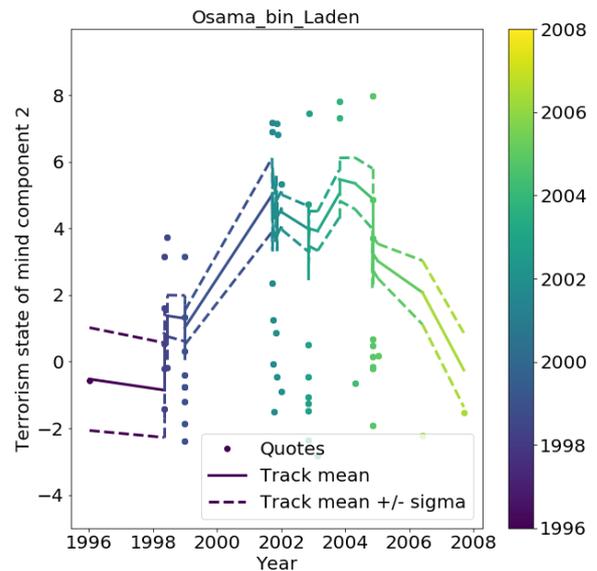

FIGURE 12.   Quote observations and state-of-mind track in dimension 2, as a function of time

## 7. Conclusions

Promising results have been obtained with an SVM quote classifier for detecting extremism or terrorism using text vector embeddings, as an alternative to n-grams, which were found to suffer from various issues. The embedding model was particularly successful in identifying terrorist quotes, but also good for detecting earlier signs of extremism. The results suggest that, for applications where it is important to distinguish signs of initial extremist tendencies from terrorism, it will be more accurate to run two separate two-way classifiers in sequence (a terrorism detector followed by an extremism detector) than to use a single three-way classifier. Further investigations are needed to understand the best way to analyse quotes of varying length, as long passages of text can be hard to classify. The tracking element of the work showed that it is possible to obtain reasonable estimates of a person's time-varying state of mind, based on the statements they make.

## 8. Acknowledgements

The work in this paper was funded by the Defence and Security Accelerator (DASA) Behavioural Analytics programme under contract DSTLX-1000134261 and the QinetiQ Fellow scheme. Internet searches and data collation were carried out using a bespoke semi-automated tool developed by Thomas Lane Cassell. © Copyright QinetiQ Limited 2020-2021.

## REFERENCES

ARLOT, S. & CELISSE, A. 2010. A survey of cross-validation procedures for model selection. *Statistics Surveys* **4**, 40–79.

BOJANOWSKI, P., GRAVE, E., JOULIN, A. & MIKOLOV, T. 2017. Enriching word vectors with subword information. *Trans. of the Assoc. for Computational Linguistics* **5**, 135–146.

BURGESS, C. J. C. 1998. A Tutorial on Support Vector Machines for Pattern Recognition. *Data Mining and Knowledge Discovery* **2**, 121–167.

CER, D., YANG, Y., KONG, S.Y., NUA, N., LIMTIACO, N., ST. JOHN, R., CONSTANT, N., GUAJARDO-CESPEDES, M., YUAN, S., TAR, C., STROPE, B. & KURZWEIL, R. 2018. Universal sentence encoder for English. *Conf. on Empirical Methods in Natural Language Processing (EMNLP), Brussels, Belgium*, pp.169–174.

DEVLIN, J., CHANG, M.W., LEE, K. & TOUTANOVA, K. 2019. BERT: Pre-training of Deep Bidirectional Transformers for Language Understanding. *arXiv:*1810.04805v2.

KOWSARI, K., MEIMANDI, K.J., HEIDARYSAFA, M., MENDU, S., BARNES, L.E. & BROWN, D.E. 2019. Text Classification Algorithms: A Survey. *arXiv:*1904.08067v4.

LANE, R.O., BRIERS, M., COOPER, T.M. & MASKELL, S.R. 2014. Efficient data structures for large scale tracking. *17th Int. Conf. on Information Fusion, Salamanca, Spain*.

LE, Q. & MIKOLOV, T. 2014. Distributed representations of sentences and documents. *Proc. Int. Conf. on Machine Learning, Beijing, China,* **32**(2):1188-1196.

MIKOLOV, T., CHEN, K., CORRADO, G. & DEAN, J. 2013. Efficient Estimation of Word Representations in Vector Space, *arXiv:*1301.3781v3.

PENNINGTON, J., SOCHER, R. & MANNING, C.D. 2014. GloVe: Global Vectors for Word Representation. *Conf. on Empirical Methods in Natural Language Processing (EMNLP), Doha, Qatar*, pp.1532–1543.

PETERS, M.E., NEUMANN, M., IYYER, M., GARDNER, M., CLARK, C., LEE, K. & ZETTLEMOYERY, L. 2018. Deep contextualized word representations. *arXiv:*1802.05365v2.

RAHMOUN, A., & ELBERRICHI, Z. 2007. Experimenting N-Grams in text categorization. *International Arab Journal of Information Technology* **4**(4):377–385.

RAMOS, J. 2003. Using tf-idf to determine word relevance in document queries. *Proc. First Instructional Conf. on Machine Learning, Piscataway, NJ, USA*, **242**, 133-142.

SPEER, R., CHIN, J. & HAVASI, C. 2017. ConceptNet 5.5: An open multilingual graph of general knowledge. *AAAI Conf. on Artificial Intelligence, San Francisco, CA, USA*.

WELCH, G. & BISHOP, G. 1995. An introduction to the Kalman filter. *University of Carolina Technical Report*.

YANG, Y., CER, D., AHMAD, A., GUO, M., LAW, J., CONSTANT, N., HERNANDEZ ABREGO, G., YUAN, S., TAR, C., SUNG, Y.H., STROPE, B. & KURZWEIL, R. 2019. Multilingual universal sentence encoder for semantic retrieval. *arXiv:1907.04307v1*.